\documentclass{bmvc2k}

\title{RL-FAT: Reinforcement Learning for Fair Adversarial Training}

\addauthor{Tejaswini Medi}{tejaswini.medi@uni-mannheim.de}{1}
\addauthor{Levan Mikeladze}{Levan.Mikeladze@iset.ge}{1}
\addauthor{Margret Keuper}{keuper@uni-mannheim.de}{1,2}

\addinstitution{\small
Chair for Machine Learning\\
University of Mannheim\\
Germany
}

\addinstitution{\small
MPI for Informatics\\
Saarland Informatics Campus\\
Germany
}

\runninghead{Medi, Mikeladze, Keuper}{RL-FAT}
\usepackage{graphicx}
\usepackage{amsmath}
\usepackage{amssymb}
\usepackage{booktabs}
\usepackage{amsmath}
\usepackage{pgfplots}
\usepackage{subcaption}
\usepackage[most]{tcolorbox}
\usepackage[table]{xcolor} 
\usepackage{xcolor}
\usepackage{booktabs}
\usepackage{array}
\usepackage{graphicx}
\usepackage{svg}
\usepackage{pgfplots}
\usepgfplotslibrary{groupplots}
\usepackage{pgfplots}
\usepgfplotslibrary{groupplots}
\usepackage{booktabs}
\usepackage{algorithm}
\usepackage{algpseudocode}
\pgfplotsset{compat=1.18}
\usepgfplotslibrary{fillbetween}
\definecolor{cleanblue}{RGB}{30,90,180}
\definecolor{robustred}{RGB}{200,45,45}
\definecolor{gapgray}{RGB}{220,220,220}

\usepackage{graphicx}
\usepackage{amsmath}
\usepackage{amssymb}
\usepackage{booktabs}
\usepackage{amsmath}
\usepackage{pgfplots}
\usepackage{subcaption}
\usepackage[most]{tcolorbox}
\usepackage[table]{xcolor} 
\usepackage{xcolor}
\usepackage{booktabs}
\usepackage{array}
\usepackage{graphicx}
\usepackage{svg}
\usepackage{wrapfig}
\usepackage{pgfplots}
\usepgfplotslibrary{groupplots}
\usepackage{pgfplots}
\usepgfplotslibrary{groupplots}
\usepackage{booktabs}
\usetikzlibrary{arrows.meta}
\pgfplotsset{compat=1.18}
\usepgfplotslibrary{fillbetween}
\definecolor{cleanblue}{RGB}{30,90,180}
\definecolor{robustred}{RGB}{200,45,45}
\definecolor{gapgray}{RGB}{220,220,220}

\begin{document}

\maketitle

\begin{abstract}

Deep neural networks remain highly vulnerable to adversarial perturbations, and adversarial training (AT) has become a widely used approach for improving robustness. However, improvements in average robust accuracy often mask substantial class-wise disparities: while some classes become more robust, others may remain disproportionately vulnerable under attack. This imbalance raises an important adversarial fairness concern, particularly in vision tasks where reliable robustness is expected across all categories. To address this challenge, we propose \textbf{RL-FAT}, a reinforcement-learning-inspired fair adversarial training framework that uses policy-gradient based feedback from adversarial predictions. RL-FAT interprets the prediction distribution as a policy and combines correctness-based prediction rewards with class-wise value estimates to compute class-specific advantages for policy-gradient optimization. This enables the model to adaptively focus on class-wise misclassification. Furthermore, we introduce a fairness-emphasis adversarial loss that assigns stronger training pressure to classes with high adversarial loss, thereby mitigating class-wise robustness disparity. By combining reinforcement-driven adaptation with fairness-emphasis regularization, RL-FAT improves adversarial robustness while promoting a more balanced robustness distribution across classes. Extensive experiments demonstrate that our method achieves competitive robust accuracy and substantially reduces class-wise robustness imbalance compared with standard adversarial training baselines.

\end{abstract}

\section{Introduction}
\label{sec:intro}

Adversarial training is one of the most effective approaches for improving the robustness of deep neural networks by training models on adversarially perturbed examples, leading to substantial gains in average robust accuracy under strong attacks~\citep{athalye2018obfuscated, Wang2020ImprovingAR, jia2022prior, grabinski2024large, Grabinskilowcut22, grabinski2022aliasing, grabinski2022robust, Jung2023, lukasik2023improving, pmlr-v235-agnihotri24b}. 
Among existing methods, TRADES~\citep{defense_trades} provides a theoretically motivated objective that balances natural accuracy and adversarial robustness, making it one of the most widely used baselines in robust optimization. 
However, improvements in average robust accuracy can still hide severe class-wise vulnerabilities~\citep{fairness_kdd, xu2021robust}. 
For example, in an autonomous driving system, a model may remain robust when recognizing common objects such as cars or road signs, while failing under small adversarial perturbations on less frequent but safety-critical classes such as pedestrians, cyclists, or emergency vehicles. 
In such cases, the model may appear reliable on average, although its behavior on hard classes determines the actual real-world risk. 
This motivates evaluating robust classifiers not only by average robust accuracy, but also by worst-class robust accuracy and class-wise robust fairness. 
Such class-wise disparities commonly studied as the \textbf{robust fairness problem}~\citep{xu2021robust, zhang2021dafa, medi2024classwiserobustnessanalysis, medi2025fair, wei2023cfa, fairness_bat, li2023wat, fairness_frl}.

\begin{figure*}[t]
\centering

\definecolor{cleanblue}{RGB}{30,90,180}
\definecolor{robustred}{RGB}{200,45,45}
\definecolor{gapgray}{RGB}{220,220,220}

{\small
\begin{tikzpicture}[baseline=-0.5ex]
    \fill[gapgray!75] (0,0) rectangle (0.35,0.16);
    \node[right] at (0.42,0.08) {Gap};

    \draw[cleanblue, thick] (1.35,0.08) -- (1.75,0.08);
    \fill[cleanblue] (1.55,0.08) circle (1.6pt);
    \node[right] at (1.82,0.08) {Clean};

    \draw[robustred, thick, dashed] (2.95,0.08) -- (3.35,0.08);
    \fill[robustred] (3.15,0.08) rectangle ++(0.08,0.08);
    \node[right] at (3.42,0.08) {Robust};

    \draw[black!65, thick, dotted] (4.75,0.08) -- (5.20,0.08);
    \node[right] at (5.27,0.08) {Mean Average};
\end{tikzpicture}
}

\vspace{1mm}

\pgfplotsset{
    bmvcaccplot/.style={
        width=\linewidth,
        height=4.25cm,
        xmin=-0.25, xmax=9.25,
        ymin=20, ymax=100,
        xtick={0,1,2,3,4,5,6,7,8,9},
        xticklabels={Air., Auto., Bird, Cat, Deer, Dog, Frog, Horse, Ship, Truck},
        xticklabel style={rotate=35, anchor=east, font=\scriptsize},
        ytick={20,40,60,80,100},
        tick label style={font=\scriptsize},
        xlabel={CIFAR-10 classes},
        ylabel={Accuracy (\%)},
        xlabel style={font=\small},
        ylabel style={font=\small},
        title style={font=\bfseries\small},
        grid=major,
        major grid style={gray!18},
        axis line style={gray!65},
        clip=false
    }
}

\begin{minipage}[t]{0.485\textwidth}
\centering
\begin{tikzpicture}
\begin{axis}[
    bmvcaccplot,
    title={(a) ResNet-18}
]

\addplot[name path=cleanR, draw=none, forget plot] coordinates {
    (0,88.8) (1,94.4) (2,76.0) (3,72.3) (4,85.1)
    (5,75.3) (6,90.9) (7,88.7) (8,94.4) (9,91.2)
};

\addplot[name path=robustR, draw=none, forget plot] coordinates {
    (0,62.6) (1,74.7) (2,35.5) (3,25.7) (4,32.0)
    (5,41.2) (6,52.1) (7,62.7) (8,70.2) (9,68.1)
};

\addplot[gapgray!75, draw=none, forget plot]
fill between[of=cleanR and robustR];

\addplot[
    cleanblue,
    thick,
    mark=*,
    mark size=1.7pt
] coordinates {
    (0,88.8) (1,94.4) (2,76.0) (3,72.3) (4,85.1)
    (5,75.3) (6,90.9) (7,88.7) (8,94.4) (9,91.2)
};

\addplot[
    robustred,
    thick,
    dashed,
    mark=square*,
    mark size=1.5pt
] coordinates {
    (0,62.6) (1,74.7) (2,35.5) (3,25.7) (4,32.0)
    (5,41.2) (6,52.1) (7,62.7) (8,70.2) (9,68.1)
};

\addplot[
    cleanblue,
    thick,
    dotted
] coordinates {
    (-0.25,85.71) (9.25,85.71)
};

\node[
    cleanblue,
    font=\scriptsize,
    anchor=west
] at (axis cs:9.05,85.71) {85.71};

\addplot[
    robustred,
    thick,
    dotted
] coordinates {
    (-0.25,52.48) (9.25,52.48)
};

\node[
    robustred,
    font=\scriptsize,
    anchor=west
] at (axis cs:9.05,52.48) {52.48};

\end{axis}
\end{tikzpicture}
\end{minipage}
\hfill
\begin{minipage}[t]{0.485\textwidth}
\centering
\begin{tikzpicture}
\begin{axis}[
    bmvcaccplot,
    title={(b) XCiT-S12},
    ylabel={}
]

\addplot[name path=cleanX, draw=none, forget plot] coordinates {
    (0,93.1) (1,96.1) (2,83.5) (3,80.4) (4,91.1)
    (5,79.2) (6,95.4) (7,92.6) (8,95.1) (9,94.1)
};

\addplot[name path=robustX, draw=none, forget plot] coordinates {
    (0,67.1) (1,75.5) (2,42.5) (3,30.4) (4,38.6)
    (5,43.4) (6,57.2) (7,66.1) (8,69.2) (9,71.3)
};

\addplot[gapgray!75, draw=none, forget plot]
fill between[of=cleanX and robustX];

\addplot[
    cleanblue,
    thick,
    mark=*,
    mark size=1.7pt
] coordinates {
    (0,93.1) (1,96.1) (2,83.5) (3,80.4) (4,91.1)
    (5,79.2) (6,95.4) (7,92.6) (8,95.1) (9,94.1)
};

\addplot[
    robustred,
    thick,
    dashed,
    mark=square*,
    mark size=1.5pt
] coordinates {
    (0,67.1) (1,75.5) (2,42.5) (3,30.4) (4,38.6)
    (5,43.4) (6,57.2) (7,66.1) (8,69.2) (9,71.3)
};

\addplot[
    cleanblue,
    thick,
    dotted
] coordinates {
    (-0.25,90.06) (9.25,90.06)
};

\node[
    cleanblue,
    font=\scriptsize,
    anchor=west
] at (axis cs:9.05,90.06) {90.06};

\addplot[
    robustred,
    thick,
    dotted
] coordinates {
    (-0.25,56.13) (9.25,56.13)
};

\node[
    robustred,
    font=\scriptsize,
    anchor=west
] at (axis cs:9.05,56.13) {56.13};

\end{axis}
\end{tikzpicture}
\end{minipage}

\vspace{0.5mm}

\caption{
Per-class clean and AutoAttack robust accuracy at $\epsilon=8/255$ on CIFAR-10.
The shaded region indicates the clean-to-robust performance gap, while dotted horizontal lines denote the corresponding mean average clean and robust accuracies.
}
\label{fig:clean_robust_gap_resnet_xcit}
\end{figure*}
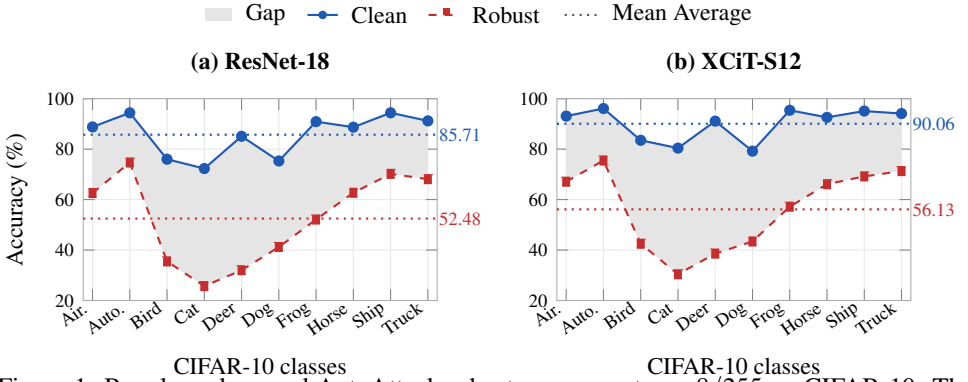

Figure~\ref{fig:clean_robust_gap_resnet_xcit} illustrates this phenomenon using ResNet-18~\cite{he_deep_2016} and XCiT-S12~\cite{elnouby2021xcit} as representative CNN-based and transformer-based RobustBench models~\cite{addepalli2022efficient, debenedetti2023light, robustbench2021}, respectively. 
For both architectures, the disparity between average and worst-class accuracy is substantially larger under adversarial evaluation than under clean evaluation. 
This suggests that robust unfairness is not limited to a single architecture family, but is a broader issue in current adversarially trained robust models. 
Therefore, improving robust fairness is necessary to ensure that robust models perform reliably across all semantic class categories.

Recent robust fairness methods aim to improve worst-class robustness while preserving competitive average robust accuracy. 
WAT~\cite{li2023wat} optimizes the worst-class objective using no-regret dynamics, but remains tied to explicit class-level worst-case updates. 
BAT~\cite{sun2021bat} shows that robust unfairness can also arise during adversarial example generation due to class-dependent attack difficulty and biased target-class tendencies, but mainly addresses these biases through designed balancing rules. 
CFA~\cite{wei2023cfa} argues that different classes prefer different adversarial training configurations, such as perturbation margins, regularization strengths, and class-wise weights, and calibrates them separately. 
DAFA~\cite{lee2024dafa} introduces an inter-class distance-aware strategy by assigning larger robustness trade-offs to visually or semantically similar classes through class-specific perturbation margins and loss weights. 
From a distributional perspective, FAAL~\cite{zhang2024faal} formulates robust fairness as a distributionally robust optimization problem and learns class-wise adversarial weights through a min-max-max objective. ABSLD~\cite{zhao2024absld} studies robust fairness in adversarial distillation by assigning class-dependent soft-label temperatures. 

Although these methods demonstrate the importance of adaptive class-aware training, they often rely on designed class-wise rules, predefined distance metrics, distillation-specific designs, or additional optimization mechanisms rather than learning fairness-aware adjustments directly from objective level training feedback.

To address this limitation, we propose \textbf{RL-FAT}, a reinforcement learning-inspired fair adversarial training framework that adaptively promotes robust fairness across classes. 
The key idea is to learn class-aware training emphasis from adversarial prediction feedback. 
In RL-FAT, the model's adversarial prediction is interpreted as an action, prediction correctness provides a positive or negative reward, and a class-wise baseline estimates the expected reward for each class. 
The resulting class-specific advantage guides policy-gradient optimization, allowing the model to focus on reducing class-wise misclassification. 
We further combine this reinforcement-driven objective with a fairness-emphasis adversarial loss, where classes with above-average adversarial loss receive stronger training emphasis. 
Together, these components enable RL-FAT to perform adaptive class-aware robust optimization without explicitly designing class-specific adversarial training configurations.

\noindent Our contributions are summarized as follows:

\begin{itemize}

    \item We propose \textbf{RL-FAT}, a reinforcement learning-inspired fair adversarial training framework that uses class-wise prediction correctness feedback, class-specific advantage estimation, and fairness-emphasis adversarial loss to improve worst-class robustness while preserving competitive average robust accuracy.

    \item We evaluate robust fairness using worst-class robust accuracy and a relative robust fairness metric that measures whether improvements in worst-class robustness exceed changes in average robustness.

    \item Experiments on CIFAR-10, CIFAR-100, and ImageNette show that RL-FAT consistently improves worst-class robust accuracy and robust fairness under strong attacks such as AutoAttack, while maintaining competitive overall clean and robust accuracy.
    
\end{itemize}

\section{Related Work}

A growing line of work studies robust fairness, where adversarially trained models can achieve high average robust accuracy while exhibiting large disparities across classes. Theoretical and empirical studies support the need to evaluate robustness beyond the average-case metric~\cite{li2023wat, sun2021bat, wei2023cfa, lee2024dafa}. 
Prior analyses show that adversarial training can amplify class-wise disparity and that the robustness--fairness trade-off becomes more severe under stronger attacks~\citep{fairness_analysis,fairness_frl}. 
Other works connect robust fairness to class imbalance and long-tailed learning, adapting class-wise reweighting strategies to adversarial training~\citep{fairness_weighting,fairness_kdd}. 
Together, these studies show that average robust accuracy alone does not capture the reliability of a robust model, and that worst-class robust accuracy improvement while maintaining the average robust accuracy is an important criterion for evaluating adversarial robust fairness.

FRL~\citep{xu2021frl} is one of the earliest methods in this direction and mitigates class-wise robust unfairness by adjusting loss weights and adversarial margins when fairness constraints are violated. 
FAT~\citep{ma2021fat} analyzes the robustness--fairness trade-off and proposes to regularize the variance of class-wise adversarial risk, motivated by the connection between adversarial risk variance and class-wise robust accuracy variance. 
BAT~\citep{sun2021bat} further studies unfairness in the adversarial example generation process and decomposes robust unfairness into source-class vulnerability, i.e., different attack difficulties across source classes, and target-class vulnerability, i.e., biased target-class tendencies of adversarial examples. 
These methods show that robust fairness requires class-aware training, but their fairness mechanisms are still specified through explicit reweighting, remargining, risk-variance regularization, or balancing rules.

More recent methods introduce richer class-wise training configurations. 
CFA~\citep{wei2023cfa} shows that different classes prefer different adversarial training configurations, including class specific perturbation margins, regularization strengths, and weight averaging, and therefore calibrates these configurations separately for each class. 
DAFA~\citep{lee2024dafa} argues that robust fairness problem is closely related to inter-class similarity: hard classes are often confused with semantically or visually similar classes rather than arbitrary easy classes. 
Based on this observation, DAFA assigns class-specific loss weights and adversarial margins using inter-class similarity distances, encouraging robustness trade-offs mainly among similar classes. 
FAAL~\citep{zhang2024faal} formulates fairness-aware adversarial learning as a distributionally robust optimization problem and introduces a min--max--max objective, where an intermediate maximization learns class-wise adversarial weights to improve worst-class robustness. 
ABSLD~\citep{zhao2024absld} studies robust fairness under adversarial robustness distillation and adjusts class-wise soft-label smoothness through different distillation temperatures. 
FAIR-TAT~\citep{medi2025fair} explores targeted adversarial training and studies its effect on fairness and robustness trade-offs across attacks and corruptions. 
Although effective, these methods instantiate fairness through predefined class-wise statistics, distance measures, temperature schedules, targeted-attack designs, or additional robust-optimization objectives.

In contrast to methods that mainly rely on predefined reweighting rules, adaptive perturbation margin strengths, class-distance estimates, temperature schedules, targeted-attack designs, or additional robust-optimization objectives, RL-FAT learns class-wise training emphasis directly from adversarial prediction feedback. During adversarial training, vulnerable classes can change over time, and improving one class may alter the robustness of others. This motivates a reinforcement-learning-inspired formulation in which the model receives correctness-based feedback from adversarial predictions and uses this signal to adapt class-wise learning during training. Related work has shown that reward-based learning for image classification can improve generalization and adversarial accuracy compared with standard cross-entropy training~\citep{gupta2020reinforcement}; however, it does not explicitly study class-wise robust fairness. Our method uses this perspective specifically for fair adversarial training: adversarial predictions are treated as actions, prediction correctness defines the reward, and class-wise reward baselines are used to form advantage-based updates. We combine this feedback signal with a fairness-emphasis adversarial loss, so that classes with above-average adversarial loss receive stronger training pressure. This enables adaptive class-aware robust optimization aimed at improving worst-class robustness while preserving competitive average robust accuracy.

\section{Methodology}

\begin{figure}
    \centering
    \includegraphics[width=\linewidth]{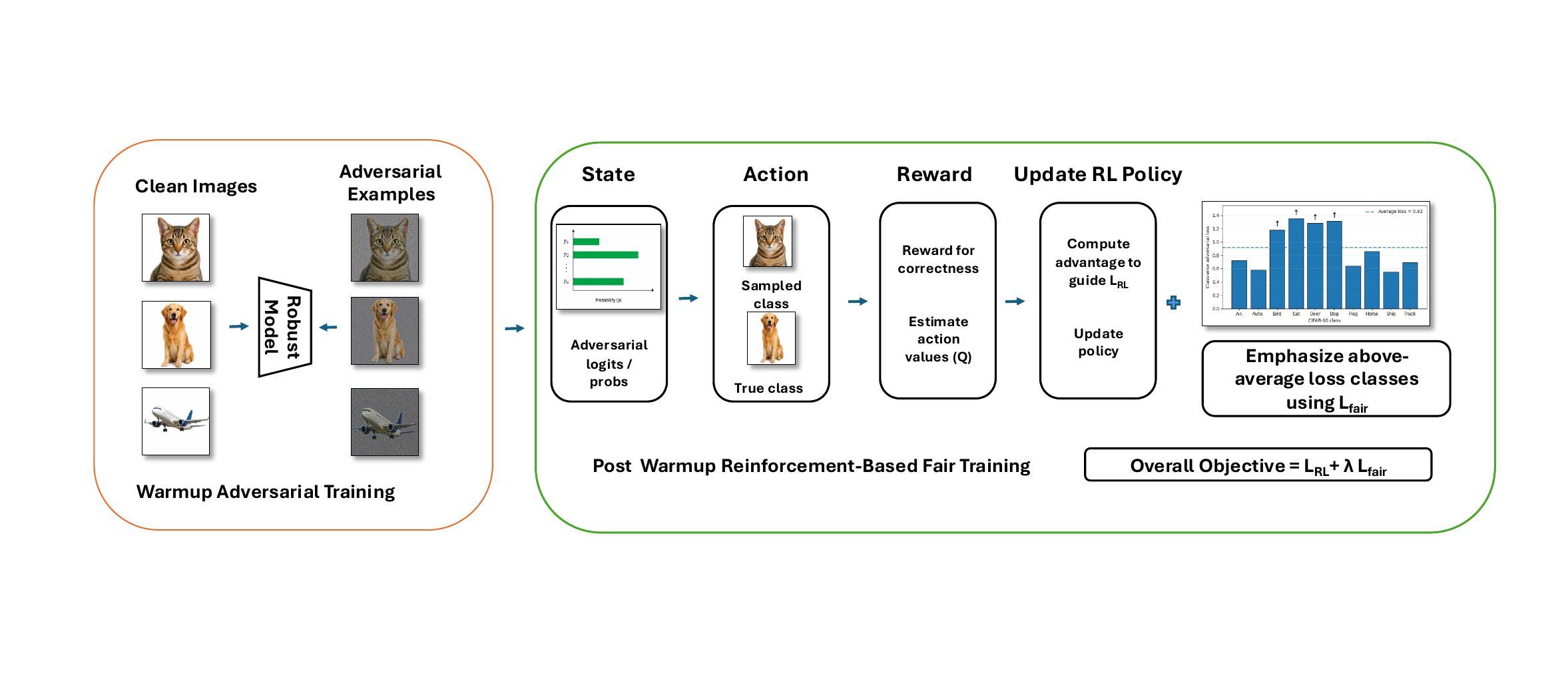}
    \caption{
    Overview of the proposed RL-FAT framework. 
    The model is first initialized using warm-up adversarial training to learn stable robust representations from clean and adversarial examples. 
    After warm-up, RL-FAT performs reinforcement-inspired fair adversarial training. 
    Given an adversarial input, the model prediction distribution is interpreted as a state, a class prediction is sampled as an action, and a reward is assigned according to prediction correctness. 
    A class-wise running value estimate is used to compute the advantage, which guides the reinforcement loss $\mathrm{L}_{\mathrm{RL}}$. 
    In parallel, the fairness-emphasis loss $\mathrm{L}_{\mathrm{fair}}$ assigns stronger training pressure to classes with above-average adversarial loss. 
    The final post-warm-up objective combines both components as $\mathrm{L}_{\mathrm{RL}} + \lambda \mathrm{L}_{\mathrm{fair}}$.
    }
    \label{fig:rlfat_overview}
\end{figure}

\subsection{Overview}

We propose \textbf{RL-FAT}, a reinforcement learning-based fair adversarial training framework for improving class-wise robust fairness. 
Given a classifier $f_{\theta}$ parameterized by $\theta$, an input image $x$, and its ground-truth label $y \in \{1,\dots,C\}$, our goal is to learn a model that is robust to adversarial perturbations while reducing robustness disparity across classes.

RL-FAT consists of two training stages. 
First, we perform a warm-up adversarial training stage using a TRADES-based robust objective~\cite{defense_trades}. 
This stage provides a stable robust initialization before applying class-wise fairness-aware optimization. 
Second, after warm-up, we optimize the model using a reinforcement-based class feedback loss together with a fairness-emphasis adversarial loss. 
The reinforcement component adaptively updates the model using reward signals derived from adversarial prediction correctness, while the fairness-emphasis component assigns stronger optimization pressure to classes whose adversarial losses are higher than the mini-batch class average. 
Unlike a hard worst-class objective that focuses only on a single class, RL-FAT can emphasize multiple vulnerable classes within each mini-batch. 
An overview of the proposed framework is shown in Figure~\ref{fig:rlfat_overview}.


\subsection{Adversarial Example Generation}
For the TRADES-based adversarial training setting, adversarial examples are generated using projected gradient descent (PGD)~\citep{madry2018towards} under the $\ell_{\infty}$ threat model,
$\|x^{adv}-x\|_{\infty}\leq\epsilon$, where $\epsilon$ is the perturbation budget. Following TRADES~\citep{defense_trades}, the adversarial example is obtained by maximizing the KL divergence between the clean and adversarial predictive distributions. Let
 $p_{\theta}(x)=\mathrm{softmax}(f_{\theta}(x))$ denote the model prediction distribution. Starting from a randomly perturbed image within the $\ell_{\infty}$ ball, PGD updates the adversarial example as:
\begin{equation}
    x^{adv}_{t+1}
    =
    \Pi_{\mathcal{B}_{\epsilon}(x)}
    \left(
    x^{adv}_{t}
    +
    \alpha \cdot
    \mathrm{sign}
    \left(
    \nabla_{x^{adv}_{t}}
    D_{\mathrm{KL}}
    \left(
    p_{\theta}(x)
    \,\|\,
    p_{\theta}(x^{adv}_{t})
    \right)
    \right)
    \right),
\end{equation}
where $\alpha$ is the PGD step size and $\Pi_{\mathcal{B}_{\epsilon}(x)}(\cdot)$ projects the perturbed image back onto the $\ell_{\infty}$ ball centered at $x$. After each step, $x^{adv}_{t+1}$ is clipped to the valid image range.

\subsection{Warm-up Robust Adversarial Training}

Directly applying reinforcement learning objectives from the beginning of training can be unstable because early predictions are noisy and class-wise feedback is unreliable. Further, Therefore, RL-FAT first performs a TRADES warm-up stage~\citep{defense_trades} by optimizing $\mathrm{L}_{\mathrm{warm}}.$ The warm-up stage provides meaningful adversarial predictions before policy optimization, avoiding extremely low-probability correct actions.
\begin{equation}
    \mathrm{L}_{\mathrm{warm}}
    =
    \mathrm{L}_{\mathrm{CE}}
    \left(
    f_{\theta}(x), y
    \right)
    +
    \beta
    D_{\mathrm{KL}}
    \left(
    p_{\theta}(x)
    \,\|\,
    p_{\theta}(x^{adv})
    \right),
\end{equation}
where $D_{\mathrm{KL}}(\cdot\|\cdot)$ is the Kullback--Leibler divergence and $\beta$ controls the robustness regularization strength. The clean cross-entropy term encourages correct prediction on natural samples, while the KL term enforces consistency between clean and adversarial predictions. After warm-up, RL-FAT switches to the proposed fairness-aware post-warm-up objective.

\subsection{Post-Warm-up Reinforcement-Based Fair Training}

After warm-up, RL-FAT optimizes the model using two complementary objectives:
\begin{equation}
    \mathrm{L}_{\mathrm{post}}
    =
    \mathrm{L}_{\mathrm{RL}}
    +
    \lambda
    \mathrm{L}_{\mathrm{fair}},
\end{equation}
where $\mathrm{L}_{\mathrm{RL}}$ is the reinforcement-based class feedback loss, $\mathrm{L}_{\mathrm{fair}}$ is the fairness-emphasis adversarial loss, and $\lambda$ controls the contribution of the fairness term. The reinforcement objective provides adaptive feedback from adversarial predictions, while the fairness-emphasis loss increases training pressure on hard classes. We detail both losses in the following. 

\subsubsection{Reinforcement-Based Class Feedback Loss}
For an adversarial input $x_i^{adv}$, the model produces a probability distribution over classes. 
We interpret this distribution as a policy:
\begin{equation}
    \pi_{\theta}(a \mid x_i^{adv})
    =
    p_{\theta}(a \mid x_i^{adv}),
\end{equation}
where $a \in \{1,\dots,C\}$ denotes a class action. 
An action is sampled from this policy $\pi_{\theta}(\cdot \mid x_i^{adv})$
and receives a reward based on whether it matches the ground-truth label:
\begin{equation}
    r_i =
    \begin{cases}
    +1, & \text{if } a_i = y_i,\\
    -1, & \text{otherwise}.
    \end{cases}
\end{equation}
Thus, correct adversarial predictions are rewarded, while incorrect adversarial predictions are penalized. 
To reduce variance and maintain class-wise feedback during training, we keep a running action-value estimate $Q_c$ for each class $c$. Let $\mathcal{I}_c = \{i : a_i = c\}$ denote the set of samples whose sampled action is class $c$. 
For every class with $|\mathcal{I}_c|>0$, the value estimate is updated using an exponential moving average:
\begin{equation}
    Q_c
    \leftarrow
    \mathrm{clip}
    \left(
    (1-\eta)Q_c
    +
    \eta
    \frac{1}{|\mathcal{I}_c|}
    \sum_{i \in \mathcal{I}_c} r_i,
    q_{\min},
    q_{\max}
    \right),
\end{equation}
where $\eta$ is the value update rate, and $q_{\min}$ and $q_{\max}$ are clipping bounds used to stabilize the value estimates. 
If $|\mathcal{I}_c|=0$, the corresponding value estimate $Q_c$ is left unchanged. For each sample, we compute the advantage as:
\begin{equation}
    A_i
    =
    \mathrm{clip}
    \left(
    r_i - Q_{c},
    A_{\min},
    A_{\max}
    \right).
\end{equation}

The class-wise value estimate $Q_c$ acts as a moving average baseline that captures the recent expected reward for actions associated with class $c$. Importantly, the index $c$ refers to the sampled action, $a_i=c$, rather than the source or ground-truth class of the input. Thus, $Q_c$ is an action-value estimate used to provide action-dependent feedback to the policy. Therefore, the advantage $A_i$ measures whether the sampled adversarial prediction performs better or worse than its current class-level expectation. The reinforcement-based class feedback loss is then defined as:
\begin{equation}
    \mathrm{L}_{\mathrm{RL}}
    =
    -
    \frac{1}{B}
    \sum_{i=1}^{B}
    \log
    \pi_{\theta}
    \left(
    a_i \mid x_i^{adv}
    \right)
    A_i,
\end{equation}
where $B$ is the mini-batch size. 
If $A_i>0$, the sampled action is reinforced by increasing its log-probability; if $A_i<0$, the sampled action is suppressed. 
This provides a dynamic class-aware feedback signal that differs from standard cross-entropy or fixed class reweighting, since the update depends on both the current adversarial prediction and the accumulated class-wise reward history. 
This objective is not intended to model a long-horizon control problem; rather, it uses policy-gradient based feedback to adapt class-wise learning signals during adversarial training. 
In this way, RL-FAT can adjust training emphasis toward samples whose adversarial prediction behavior remains unreliable over training time.

\subsubsection{Fairness-Emphasis Adversarial Loss}

Although the reinforcement objective provides adaptive prediction-level feedback, it does not explicitly ensure that hard classes receive stronger adversarial training emphasis. 
To address this, we introduce a fairness-emphasis loss that identifies classes with above-average adversarial loss within each mini-batch. Let $\mathcal{B}_c = \{i : y_i = c\}$ be the set of mini-batch samples belonging to class $c$. For each class present in the mini-batch, we compute the class-wise adversarial cross-entropy loss:
\begin{equation}
    \mathrm{L}_c
    =
    \frac{1}{|\mathcal{B}_c|}
    \sum_{i \in \mathcal{B}_c}
    \mathrm{L}_{\mathrm{CE}}
    \left(
    f_{\theta}(x_i^{adv}), y_i
    \right).
\end{equation}
Let $\mathcal{C}_{B}=\{c:|\mathcal{B}_c|>0\}$ denote the set of classes appearing in the mini-batch. 
The average class-wise adversarial loss is defined as:
\begin{equation}
    \overline{\mathrm{L}}
    =
    \frac{1}{|\mathcal{C}_{B}|}
    \sum_{c \in \mathcal{C}_{B}}
    \mathrm{L}_c.
\end{equation}
A class is considered hard if its adversarial loss is larger than the mini-batch class average. 
We define the hard-class gap as:
\begin{equation}
    g_c
    =
    \max
    \left(
    \mathrm{L}_c
    -
    \overline{\mathrm{L}},
    0
    \right).
\end{equation}
The hard-class weights are obtained by normalizing the positive gaps
\begin{equation}
    w_c
    =
    \frac{g_c}
    {
    \sum_{j \in \mathcal{C}_{B}} g_j 
    },
\end{equation}
The weights are detached from the computational graph so that gradients optimize the adversarial classification loss rather than the weighting function itself. 
The fairness-emphasis loss is defined as $\mathrm{L}_{\text{fair}}$. This fairness-emphasis loss differs from a hard worst-class objective. 
Instead of selecting only the single class with the maximum adversarial loss, it distributes training pressure across all classes that are currently weaker than the mini-batch average. 
This provides a smoother fairness signal and allows multiple vulnerable classes to be improved simultaneously. 
Overall, the reinforcement loss encourages adaptive correction of adversarial prediction behavior, while the fairness-emphasis loss directly targets hard classes with above-average adversarial loss. 
Together, these components improve class-wise robust fairness without relying only on average robust accuracy.

\begin{equation}
    \mathrm{L}_{\text{fair}}
    =
    \sum_{c \in \mathcal{C}_{B}}
    w_c
    \mathrm{L}_c.
\end{equation}


\subsection{Class-wise Robust Fairness Evaluation}

To evaluate robust fairness, we measure adversarial robustness at the class level. 
For class $c$, the class-wise robust accuracy is defined as:
\begin{equation}
    A_c^{rob}(f)
    =
    \frac{1}{N_c}
    \sum_{i:y_i=c}
    \mathbf{1}
    \left[
    \arg\max_j f_{\theta}(x_i^{adv})_j = y_i
    \right],
\end{equation}
where $N_c$ is the number of test samples in class $c$, and $x_i^{adv}$ is the adversarial example. 
We report both the average robust accuracy $A_{avg}^{rob}(f)$ and the worst-class robust accuracy, which measures the robustness of the most vulnerable class.
\begin{equation}
    A_{worst}^{rob}(f)
    =
    \min_c A_c^{rob}(f),
\end{equation}

Following the robust fairness motivation in DAFA~\cite{lee2024dafa}, we further define a relative robust fairness metric with respect to a baseline model $f_0$:
\begin{equation}
    \rho_{\mathrm{rob}}(f; f_0)
    =
    100
    \left[
    \frac{
    A_{worst}^{rob}(f)-A_{worst}^{rob}(f_0)
    }{
    A_{worst}^{rob}(f_0)
    }
    -
    \frac{
    A_{avg}^{rob}(f)-A_{avg}^{rob}(f_0)
    }{
    A_{avg}^{rob}(f_0)
    }
    \right],
\end{equation}
This metric measures whether the improvement in worst-class robust accuracy exceeds the corresponding change in average robust accuracy. Also, this metric is baseline-dependent and captures fairness only at the class level, so it may not fully reflect subgroup robust fairness. A larger $\rho_{\mathrm{rob}}$ indicates stronger robust fairness improvement. 
Thus, RL-FAT aims to increase $A_{worst}^{rob}$ while maintaining strong $A_{avg}^{rob}$, leading to improved robust fairness.

\section{Experiments}

In this section, we detail the efficacy of our approach RL-FAT utilizing the vision benchmark datasets and models in adversarial fairness domain.\label{sec:experiments}

\subsection{Datasets and Baselines}
We evaluate the proposed RL-FAT framework on standard image classification benchmarks commonly used in adversarial robustness and robust fairness studies. 
The evaluation includes CIFAR-10, CIFAR-100~\cite{cifar10}, and ImageNette~\cite{imagenette, imagenet}. We use TRADES as the main adversarial training baseline and compare against representative robust fairness methods, including \textbf{FRL}~\cite{xu2021frl}, \textbf{WAT}~\cite{li2023wat}, \textbf{CFA}~\cite{wei2023cfa}, \textbf{DAFA}~\cite{lee2024dafa}. 
All methods are trained using PGD and evaluated with AutoAttack~\cite{lorenz2022is} under the same perturbation budget and model architecture opted are CNN based architecture ResNet-18~\cite{he_deep_2016} and we also include Transformer based architecture XCiT-S12~\cite{elnouby2021xcit} to show generalization. We train our approach and DAFA for 150 epochs, while methods with longer schedules follow their official implementations for fair comparison.

\subsection{Implementation Details and Metrics}

We use ResNet-18~\cite{he_deep_2016} as the main backbone and train all models with SGD using momentum $0.9$ and weight decay $2\times10^{-4}$. All experiments follow the $\ell_{\infty}$ adversarial threat model with perturbation budget $\epsilon=8/255$. We adopt a two-stage training procedure: first, we train with the TRADES objective~\cite{defense_trades} using $\beta=6$ for $110$ epochs to obtain a stable robust initialization; then, we continue training from this warm-up checkpoint with the proposed RL-FAT objective until epoch $150$. During the TRADES warm-up stage, the initial learning rate is set to $0.1$ and decayed by factors of $10$ and $100$ in the last $10$ and last $5$ epochs, respectively. We use a training batch size of $256$ and a test batch size of $100$. Adversarial examples are generated using $20$-step PGD with step size $0.00784$ and are clipped to valid image range.

In the RL-FAT stage, for each adversarial mini-batch, class actions are sampled from the model prediction probabilities and assigned rewards of $+1$ for correct actions and $-1$ otherwise. The class-wise action-value estimate is updated with learning rate $0.05$ and clipped to $[q_{\min},q_{\max}]=[-0.5,1.0]$, while the advantage is clipped to $[A_{\min},A_{\max}]=[-2.0,2.0]$. The RL-FAT objective combines the reinforcement loss with a fairness loss that gives higher weight to classes with above-average adversarial loss. 

All experiments are run on RTX A6000 GPUs with $47.55$ GB VRAM and repeated over five random seeds. Checkpoints are saved during training, and for each seed we select one checkpoint using only the training-time validation criterion that balances average robust accuracy and worst-class robust accuracy. The test set is not used for checkpoint selection. Final results are reported by evaluating the selected checkpoint once with AutoAttack under the same perturbation budget $\epsilon=8/255$. RL-FAT introduces only lightweight class-wise value updates with modest computational overhead; training time and memory costs are reported in Supplementary~\ref{CC}.


\subsection{Evaluations on CIFAR-10}
Table~\ref{tab:cifar10_main_resnet} compares RL-FAT with standard adversarial training and existing robust fairness methods on CIFAR-10 using a ResNet-18 backbone under AutoAttack evaluation. The goal is to evaluate whether RL-FAT improves worst-class robust accuracy while preserving competitive average robust accuracy. RL-FAT achieves the highest worst-class robust accuracy of $32.38\%$, improving over TRADES by $8.58\%$ . Compared with robust fairness baselines, RL-FAT improves worst-class robust accuracy by $4.78\%$, $2.94\%$, $5.17\%$, and $1.36\%$  over FRL, WAT, CFA, and DAFA, respectively. RL-FAT also obtains the highest average clean accuracy of $82.99\%$ and the best $\rho_{\mathrm{rob}}$ score of $38.77$, indicating the strongest robust fairness improvement among all methods. Although RL-FAT does not achieve the highest average robust accuracy, its average robust accuracy of $47.88\%$ remains competitive, while it provides the largest gain for the most vulnerable class. These results show that RL-FAT achieves a better adversarial fairness trade-off by improving worst-class robustness rather than only maximizing average robust accuracy.
\begin{table*}[t]
\small
\centering
\caption{
Robust fairness comparison on CIFAR-10. 
We report average and worst-class accuracy under clean and adversarial evaluation for the best checkpoint evaluations using AutoAttack~\cite{lorenz2022is} with epsilon $\epsilon=8/255$ on ResNet-18 over five seeds. 
The fairness metric $\rho_{\mathrm{rob}}$ measures the relative improvement in worst-class robust accuracy over the TRADES baseline, adjusted by the relative change in average robust accuracy.
Higher $\rho_{\mathrm{rob}}$ indicates better robust fairness improvement.
The best result in each column is set in bold.
}
\vspace{1.5mm}
\label{tab:cifar10_main_resnet}
\begin{tabular}{@{}lccccc@{}}
\toprule
\textbf{Method} 
& \textbf{Avg. Clean} 
& \textbf{Worst Clean} 
& \textbf{Avg. Robust} 
& \textbf{Worst Robust} 
& \textbf{ $\rho_{\mathrm{rob}}$} \\
\midrule
TRADES        & 81.75 $\pm$ 0.61 & 66.52 $\pm$ 1.53 & $49.22 \pm 0.74$ & 23.80 $\pm$ 1.35 & 0.00 \\
FRL           & 82.02 $\pm$ 0.20 & \textbf{70.27 $\pm$ 0.76} & 44.91 $\pm$ 0.30 & 27.60 $\pm$ 0.74 & 24.72 \\
WAT           & 79.02 $\pm$ 0.21 & 70.00 $\pm$ 1.31 & 45.92 $\pm$ 0.19 & 29.44 $\pm$ 1.39 & 30.40 \\
CFA           & 79.02 $\pm$ 0.13 & 64.01 $\pm$ 0.04 & \textbf{49.30 $\pm$ 0.21} & 27.21 $\pm$ 0.67 & 14.17 \\
DAFA          & 81.49 $\pm$ 0.63 & 66.66 $\pm$ 0.66 & 48.03 $\pm$ 0.80 & 31.02 $\pm$ 0.66 & 32.75 \\
\textbf{RL-FAT} &  \textbf{82.99 $\pm$ 0.09} & 67.52 $\pm$ 0.44 & 47.88 $\pm$ 0.20 & \textbf{32.38 $\pm$ 0.44} & \textbf{38.77} \\
\bottomrule
\end{tabular}
\end{table*}

\vspace{-7mm}

\subsection{Evaluations on CIFAR-100}
\begin{table*}[t]
\small
\centering
\caption{
Robust fairness comparison on CIFAR-100. We report mean and standard deviation over five seeds for the best checkpoint evaluations using AutoAttack on ResNet-18 with $\epsilon=8/255$. 
The fairness metric $\rho_{\mathrm{rob}}$ measures the relative improvement in worst-class robust accuracy over the TRADES baseline, adjusted by the relative change in average robust accuracy.
Higher $\rho_{\mathrm{rob}}$ indicates better robust fairness improvement.
}
\vspace{2mm}
\label{tab:cifar100_main}
\setlength{\tabcolsep}{5.5pt}
\renewcommand{\arraystretch}{1.08}
\begin{tabular}{@{}lccccc@{}}
\toprule
\textbf{Method} 
& \textbf{Avg. Clean} 
& \textbf{Worst Clean} 
& \textbf{Avg. Robust} 
& \textbf{Worst Robust} 
& \textbf{ $\rho_{\mathrm{rob}}$} \\
\midrule
TRADES 
& $58.12 \pm 0.82$ 
& $16.80 \pm 2.17$ 
& $25.13 \pm 0.36$ 
& $1.30 \pm 0.84$ 
& $0.00$ \\

FRL 
& $\mathbf{59.03 \pm 1.02}$ 
& $\mathbf{26.00 \pm 3.32}$ 
& $21.97 \pm 0.49$ 
& $1.40 \pm 0.89$ 
& $20.27$ \\

WAT 
& $45.34 \pm 0.71$ 
& $19.60 \pm 1.52$ 
& $16.01 \pm 0.47$ 
& $1.60 \pm 0.84$ 
& $59.37$ \\

CFA 
& $58.25 \pm 0.82$ 
& $18.20 \pm 2.17$ 
& $\mathbf{29.44 \pm 0.32}$ 
& $1.80 \pm 0.84$ 
& $21.31$ \\

DAFA 
& $57.94 \pm 0.53$ 
& $19.80 \pm 2.17$ 
& $24.42 \pm 0.58$ 
& $2.10 \pm 1.64$ 
& $64.36$ \\

\textbf{RL-FAT} 
& $56.64 \pm 0.24$ 
& $18.80 \pm 1.30$ 
& $24.81 \pm 0.15$ 
& $\mathbf{2.20 \pm 0.45}$ 
& $\mathbf{70.50}$ \\
\bottomrule
\end{tabular}
\end{table*}

Table~\ref{tab:cifar100_main} compares RL-FAT with standard adversarial training and existing robust fairness methods on CIFAR-100 using an adversarially trained ResNet-18 model under AutoAttack evaluation. This setting evaluates whether RL-FAT can improve worst-class robust accuracy when the number of classes is substantially larger than CIFAR-10. As shown in the table, RL-FAT achieves the best worst-class robust accuracy of $2.20\%$, outperforming TRADES, FRL, WAT, CFA, and DAFA. Compared with TRADES, RL-FAT improves worst-class robust accuracy by $0.90\%$, while also obtaining a competitive average robust accuracy of $24.81\%$. Although CFA achieves the highest average robust accuracy of $29.44\%$, RL-FAT provides the strongest robust fairness improvement, achieving the best $\rho_{\mathrm{rob}}$ score of $70.50$. These results indicate that RL-FAT improves robustness for the most vulnerable classes and remains effective in the more number of classes of CIFAR-100 setting, where class-wise robustness disparity is harder to control due to the larger number of hard classes.

\subsection{Evaluations on ImageNette}

\begin{table*}[t]
\small
\centering
\caption{
Robust fairness comparison on ImageNette. 
We report average and worst-class accuracy under clean and adversarial evaluation, with values shown as mean and standard deviation over five seeds. 
Adversarial robustness is evaluated using AutoAttack with ResNet-18. we evaluate the robust fairness metric $\rho_{\mathrm{rob}}$, computed relative to the TRADES baseline. Higher values indicate stronger robust fairness improvement. Best results are set in bold.
}
\vspace{1mm}
\label{tab:Imagenette_main}
\begin{tabular}{@{}lccccc@{}}
\toprule
\textbf{Method} 
& \textbf{Avg. Clean} 
& \textbf{Worst Clean} 
& \textbf{Avg. Robust} 
& \textbf{Worst Robust} 
& \textbf{$\rho_{\mathrm{rob}}\uparrow$} \\
\midrule
TRADES       
& $76.00 \pm 0.83$ 
& $58.74 \pm 0.88$ 
& $48.66 \pm 0.38$ 
& $22.90 \pm 1.18$ 
& $0.00$ \\

FRL          
& $\mathbf{76.76 \pm 1.38}$ 
& $\mathbf{64.34 \pm 1.42}$ 
& $45.75 \pm 1.48$ 
& $27.35 \pm 3.04$ 
& $25.41$ \\

WAT          
& $50.24 \pm 1.20$ 
& $39.77 \pm 2.27$ 
& $24.13 \pm 0.34$ 
& $11.26 \pm 1.90$ 
& $-0.42$ \\

CFA          
& $75.46 \pm 0.25$ 
& $51.23 \pm 2.52$ 
& $\mathbf{50.68 \pm 0.86}$ 
& $23.50 \pm 1.79$ 
& $-1.53$ \\

DAFA         
& $75.81 \pm 0.29$ 
& $63.89 \pm 1.56$ 
& $48.17 \pm 0.63$ 
& $28.08 \pm 1.53$ 
& $23.63$ \\

\textbf{RL-FAT} 
& $75.92 \pm 0.31$ 
& $61.96 \pm 0.87$ 
& $47.91 \pm 0.52$ 
& $\mathbf{28.79 \pm 1.77}$ 
& $\mathbf{27.26}$ \\
\bottomrule
\end{tabular}
\end{table*}

Table~\ref{tab:Imagenette_main} reports the robust fairness results on ImageNette, a higher-resolution dataset than CIFAR-10. This setting evaluates whether RL-FAT can maintain class-wise robust fairness when the input resolution increases. We evaluate ImageNette at $64 \times 64$ resolution. As shown in the table, RL-FAT achieves the best worst-class robust accuracy of $28.79\%$, outperforming TRADES, FRL, WAT, CFA, and DAFA. Compared with TRADES, RL-FAT improves worst-class robust accuracy by $5.89\%$, showing that it provides stronger protection for the most vulnerable class under adversarial perturbations.

RL-FAT also maintains competitive overall performance, achieving an average clean accuracy of $75.92\%$ and an average robust accuracy of $47.91\%$. Although CFA obtains the highest average robust accuracy of $50.68\%$, its worst-class robust accuracy is lower than that of RL-FAT. Similarly, while FRL achieves the best average clean and worst-class clean accuracy, RL-FAT provides stronger worst-class robustness under adversarial evaluation. RL-FAT also achieves the highest $\rho_{\mathrm{rob}}$ score of $27.26$, indicating the strongest robust fairness improvement among all methods. Overall, these results show that RL-FAT remains effective in the higher-resolution ImageNette setting by improving the robustness of the most vulnerable class without severely sacrificing average clean or robust accuracy.

\subsection{Class-specific Robustness Steering}
We further analyze whether RL-FAT can be used to steer robustness toward a user-specified class. 
This experiment is not intended to improve robust fairness directly; instead, it evaluates the controllability of the RL-FAT training signal and its effect on target-class, average, and worst-class performance. Let $t$ denote a selected target class. 
To steer training toward target class $t$, we replace the fairness-emphasis loss with the adversarial loss of the target class, $ \mathrm{L}_{steer}(t) = \mathrm{L}_{t}$, where $\mathrm{L}_{t}$ is the adversarial cross-entropy loss for samples belonging to class $t$ in the mini-batch. 
The post warm-up objective is then given by:
\begin{equation}
    \mathrm{L}_{post}(t)
    =
    \mathrm{L}_{\mathrm{RL}}
    +
    \lambda
    \mathrm{L}_{steer}(t).
\end{equation}
By changing $t$, RL-FAT can redirect training pressure toward different class-specific robustness goals. Table~\ref{tab:class_specific_steering} reports the results of steering RL-FAT toward each CIFAR-10 class. 
The target-class clean and robust accuracies measure the direct effect of steering on the selected class, while the average and worst-class accuracies show how this targeted emphasis affects the overall performance distribution. 
The results show that steering can substantially improve the selected target class. 
For example, steering toward Cat achieves the highest target clean accuracy of $99.72\%$ and the highest target robust accuracy of $94.42\%$, while steering toward Frog, Dog, Deer, and Bird also yields target robust accuracies above $90\%$. 
This confirms that RL-FAT can intentionally allocate robustness pressure toward a desired class.

At the same time, steering toward a single class can shift robustness across other classes. For instance, steering toward Horse achieves the highest average clean accuracy of $76.82\%$, average robust accuracy of $47.32\%$, and worst-class robust accuracy of $16.74\%$, while Truck yields the strongest worst-class clean accuracy of $36.12\%$. These results show that the target class affects both its own performance and the overall class-wise robustness distribution.

\begin{table*}[t]
\small
\centering
\small
\caption{
Class-specific performance steering on CIFAR-10 using a ResNet-18 model.
Each row reports results when RL-FAT is steered toward the target class listed in the first column.
We report the average accuracy across all classes, the worst-class accuracy, and the accuracy of the steered target class under both clean evaluation and adversarial evaluation with AutoAttack.
For the worst-class columns, we also report the class that most frequently attains the lowest accuracy across five random seeds; the value in parentheses indicates how many of the five seeds identified that class as the worst-performing class.
}
\vspace{1mm}
\label{tab:class_specific_steering}
\resizebox{\textwidth}{!}{
\begin{tabular}{@{}lcccccccc@{}}
\toprule
\textbf{Steered Class} 
& \textbf{Avg. Clean} 
& \textbf{Worst Clean} 
& \textbf{Worst Clean Class}
& \textbf{Target Clean} 
& \textbf{Avg. Robust} 
& \textbf{Worst Robust} 
& \textbf{Worst Robust Class}
& \textbf{Target Robust} \\
\midrule
Airplane 
& $71.20 \pm 0.50$ 
& $32.26 \pm 0.86$ 
& Dog (4/5) 
& $96.12 \pm 0.13$ 
& $43.90 \pm 0.31$ 
& $13.90 \pm 0.46$ 
& Dog (3/5) 
& $79.18 \pm 0.65$ \\

Automobile 
& $70.77 \pm 0.55$ 
& $33.20 \pm 0.87$ 
& Dog (5/5) 
& $97.52 \pm 0.49$ 
& $44.80 \pm 0.24$ 
& $14.68 \pm 0.54$ 
& Dog (4/5) 
& $80.88 \pm 1.47$ \\

Bird 
& $60.70 \pm 0.76$ 
& $14.46 \pm 2.15$ 
& Deer (5/5) 
& $99.56 \pm 0.11$ 
& $40.01 \pm 0.65$ 
& $4.64 \pm 0.45$ 
& Deer (5/5) 
& $91.84 \pm 1.58$ \\

Cat 
& $59.01 \pm 0.28$ 
& $17.88 \pm 1.29$ 
& Dog (4/5) 
& $\mathbf{99.72 \pm 0.11}$ 
& $38.97 \pm 0.26$ 
& $5.14 \pm 0.93$ 
& Deer (5/5) 
& $\mathbf{94.42 \pm 0.61}$ \\

Deer 
& $64.41 \pm 0.68$ 
& $28.14 \pm 1.07$ 
& Bird (5/5) 
& $99.56 \pm 0.18$ 
& $42.78 \pm 0.38$ 
& $11.86 \pm 0.86$ 
& Frog (5/5) 
& $92.98 \pm 0.85$ \\

Dog 
& $61.03 \pm 0.63$ 
& $1.30 \pm 1.05$ 
& Cat (5/5) 
& $99.60 \pm 0.10$ 
& $40.36 \pm 0.31$ 
& $0.26 \pm 0.30$ 
& Cat (5/5) 
& $93.24 \pm 0.42$ \\

Frog 
& $69.04 \pm 0.57$ 
& $30.96 \pm 2.47$ 
& Deer (5/5) 
& $99.64 \pm 0.11$ 
& $45.75 \pm 0.22$ 
& $8.98 \pm 0.94$ 
& Deer (5/5) 
& $93.86 \pm 0.94$ \\

Horse 
& $\mathbf{76.82 \pm 0.36}$ 
& $35.60 \pm 0.91$ 
& Dog (5/5) 
& $93.96 \pm 0.43$ 
& $\mathbf{47.32 \pm 0.18}$ 
& $\mathbf{16.74 \pm 1.05}$ 
& Dog (5/5) 
& $71.20 \pm 0.66$ \\

Ship 
& $70.34 \pm 0.50$ 
& $33.54 \pm 0.67$ 
& Dog (5/5) 
& $97.62 \pm 0.22$ 
& $45.17 \pm 0.30$ 
& $14.96 \pm 0.45$ 
& Dog (4/5) 
& $82.74 \pm 1.17$ \\

Truck 
& $70.78 \pm 0.56$ 
& $\mathbf{36.12 \pm 1.45}$ 
& Bird (4/5) 
& $96.66 \pm 0.21$ 
& $45.15 \pm 0.16$ 
& $15.94 \pm 1.03$ 
& Frog (3/5) 
& $80.08 \pm 1.29$ \\
\bottomrule
\end{tabular}
}
\end{table*}

\vspace{-3mm}

\subsection{Evaluation on Out-of-Distribution Detection}
\label{sec:ood_svhn}

\begin{figure*}[t]
\centering
\small
\begin{tikzpicture}
\begin{groupplot}[
group style={
group size=2 by 1,
horizontal sep=0.9cm
},
width=0.40\textwidth,
height=4.8cm,
xmin=0.5,
xmax=5.5,
ymin=0,
ymax=100,
ylabel={Score},
xtick={1,2,3,4,5},
xticklabels={TRADES,FRL,WAT,DAFA,RL-FAT},
x tick label style={rotate=35, anchor=east, font=\scriptsize},
y tick label style={font=\scriptsize},
ymajorgrids=true,
grid style=dashed,
tick label style={font=\scriptsize},
label style={font=\scriptsize},
title style={font=\small},
legend style={
at={(1.05,1.22)},
anchor=south,
legend columns=3,
draw=none,
font=\scriptsize,
/tikz/every even column/.append style={column sep=0.25cm}
}
]

\nextgroupplot[
title={Overall OOD Detection}
]

\addplot[
only marks,
mark=*,
mark size=2.2pt,
blue
] coordinates {
(1,79.05)
(2,85.90)
(3,64.68)
(4,78.82)
(5,83.78)
};
\addlegendentry{AUROC}

\addplot[
only marks,
mark=square*,
mark size=2.2pt,
red
] coordinates {
(1,69.27)
(2,79.01)
(3,50.14)
(4,71.51)
(5,78.57)
};
\addlegendentry{AUPR-IN}

\addplot[
only marks,
mark=triangle*,
mark size=2.5pt,
brown!80!black
] coordinates {
(1,86.56)
(2,91.16)
(3,76.25)
(4,85.04)
(5,88.65)
};
\addlegendentry{AUPR-OUT}

\addplot[
only marks,
mark=o,
mark size=4.3pt,
line width=0.8pt,
green!50!black,
forget plot
] coordinates {
(5,83.78)
(5,78.57)
(5,88.65)
};

\nextgroupplot[
title={Worst-Class MaxLogit Metrics},
ylabel={}
]

\addplot[
only marks,
mark=*,
mark size=2.2pt,
blue,
forget plot
] coordinates {
(1,65.84)
(2,76.58)
(3,46.98)
(4,67.35)
(5,77.79)
};

\addplot[
only marks,
mark=square*,
mark size=2.2pt,
red,
forget plot
] coordinates {
(1,7.59)
(2,12.83)
(3,4.46)
(4,8.78)
(5,24.86)
};

\addplot[
only marks,
mark=triangle*,
mark size=2.5pt,
brown!80!black,
forget plot
] coordinates {
(1,97.25)
(2,98.36)
(3,94.44)
(4,97.30)
(5,98.40)
};

\addplot[
only marks,
mark=o,
mark size=4.3pt,
line width=0.8pt,
green!50!black,
forget plot
] coordinates {
(5,77.79)
(5,24.86)
(5,98.40)
};

\end{groupplot}
\end{tikzpicture}

\caption{
SVHN out-of-distribution detection performance with CIFAR-10 as the in-distribution dataset.
The left plot shows overall MaxLogit metrics, while the right plot shows worst-class MaxLogit metrics.
Green circles highlight RL-FAT results.
Higher values indicate better performance.
}
\label{fig:svhn_ood_dotplot}

\end{figure*}
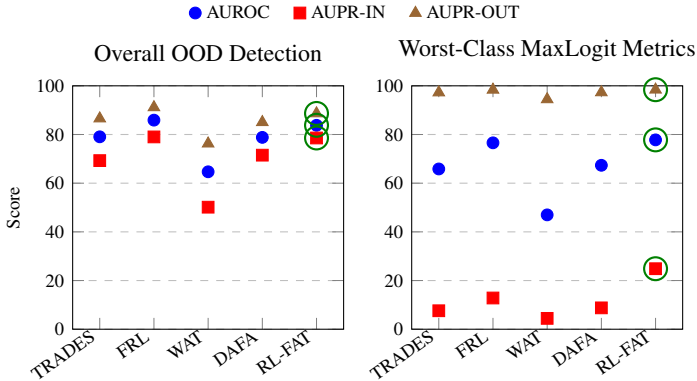

We further evaluate whether robust fairness transfers to out-of-distribution (OOD) detection. OOD detection is relevant because a robust and fair model should not only perform evenly across in-distribution classes under adversarial perturbations, but also reliably separate unfamiliar inputs from all classes without leaving particular classes more vulnerable. Following standard OOD protocols~\citep{hendrycks2017baseline,liang2018odin,yang2022openood}, we use CIFAR-10~\citep{krizhevsky2009learning} as the in-distribution dataset and SVHN~\citep{netzer2011reading} as the OOD dataset, which is semantically different from CIFAR-10 object categories. We use MaxLogit as the OOD score, assigning each image the maximum pre-softmax logit over CIFAR-10 classes~\citep{zhang2023decoupling}, and report AUROC, AUPR-IN, and AUPR-OUT~\citep{hendrycks2017baseline,liang2018odin,yang2022openood}. In addition to overall performance, we compute worst-class OOD performance by evaluating each CIFAR-10 class against all SVHN samples and taking the worst value across classes. As shown in Figure~\ref{fig:svhn_ood_dotplot}, RL-FAT achieves the best worst-class AUROC, AUPR-IN, and AUPR-OUT among the evaluated methods, indicating improved class-balanced OOD separation while maintaining competitive overall OOD detection performance.

\subsection{Ablation on RL-FAT}
\label{sec:ablation}

Table~\ref{tab:ablation_rlfat} presents an ablation study of the main RL-FAT components on CIFAR-10 under AutoAttack at $\epsilon=8/255$. TRADES achieves the highest average robust accuracy of $49.22\%$, but it has the lowest worst-class robust accuracy of $23.80\%$, indicating limited robust fairness improvement. Removing the fairness loss improves average clean accuracy to $83.06\%$, but its worst-class robust accuracy remains lower at $27.00\%$, showing that the fairness loss is important for improving the most vulnerable class. The worst-case loss variant further improves worst-class robust accuracy to $31.60\%$ and achieves a stronger $\rho_{\mathrm{rob}}$ score of $35.74$. However, the proposed RL-FAT objective achieves the best worst-class robust accuracy of $32.38\%$ and the highest $\rho_{\mathrm{rob}}$ score of $38.77$. These results demonstrate that the complete RL-FAT objective provides the strongest robust fairness improvement by enhancing worst-class robustness while maintaining competitive average clean and robust accuracy.

\begin{table*}[t]
\centering
\caption{
Ablation study of RL-FAT components on CIFAR-10. 
All variants are evaluated using AutoAttack at $\epsilon=8/255$ using ResNet-18. 
We report average and worst-class accuracy under clean and adversarial evaluation along with fairness metric $\rho_{\mathrm{rob}}$.  
Higher clean accuracy, robust accuracy, worst-class accuracy, and $\rho_{\mathrm{rob}}$ are better.
The best result in each column is set in bold.
}
\vspace{1.5mm}
\label{tab:ablation_rlfat}
\resizebox{\textwidth}{!}{
\begin{tabular}{lccccc}
\toprule
\textbf{Variant}
& \textbf{Avg. Clean} $\uparrow$
& \textbf{Worst Clean} $\uparrow$
& \textbf{Avg. Robust} $\uparrow$
& \textbf{Worst Robust} $\uparrow$
& \textbf{$\rho_{\mathrm{rob}}$} $\uparrow$ \\
\midrule
TRADES        
& $81.75 \pm 0.61$ 
& $66.52 \pm 1.53$ 
& $\textbf{49.22} \pm \textbf{0.74}$ 
& $23.80 \pm 1.35$ 
& $0.00$ \\

RL-FAT w/o fairness loss     
& $\textbf{83.06} \pm \textbf{0.27}$ 
& $67.06 \pm 0.87$ 
& $48.88 \pm 0.14$ 
& $27.00 \pm 0.67$ 
& $14.14$ \\

RL-FAT worst-case loss      
& $83.01 \pm 0.35$ 
& $\textbf{67.77} \pm \textbf{2.01}$ 
& $47.76 \pm 0.43$ 
& $31.60 \pm 0.53$ 
& $35.74$ \\

\textbf{RL-FAT} 
& $82.99 \pm 0.09$ 
& $67.52 \pm 0.44$ 
& $47.88 \pm 0.20$ 
& $\textbf{32.38} \pm \textbf{0.44}$ 
& $\textbf{38.77}$ \\
\bottomrule
\end{tabular}
}
\end{table*}

\subsection{Ablation on PGD-Based Warm-up }
\label{sec:ablation_pgd}

\begin{table*}[t]
\centering
\small
\caption{
Robust fairness comparison on CIFAR-10 using PGD adversarial training on ResNet-18.
We report average and worst-class accuracy under clean and adversarial evaluation for the best checkpoint evaluations using AutoAttack~\cite{lorenz2022is} with epsilon $\epsilon=8/255$.
The best result in each accuracy column is highlighted in bold, while the highest fairness metric is highlighted in bold.
}
\vspace{1mm}
\label{tab:cifar10_main_pgd}
\begin{tabular}{@{}lccccc@{}}
\toprule
\textbf{Method}
& \textbf{Avg. Clean}
& \textbf{Worst Clean}
& \textbf{Avg. Robust}
& \textbf{Worst Robust}
& \textbf{$\rho_{\mathrm{rob}}\uparrow$} \\
\midrule
PGD
& $84.06 \pm 0.65$
& $64.58 \pm 1.25$
& $48.80 \pm 0.38$
& $19.24 \pm 1.06$
& $0.00$ \\

FRL
& $83.53 \pm 0.31$
& $64.88 \pm 2.01$
& $45.54 \pm 0.37$
& $23.16 \pm 1.30$
& $27.05$ \\

CFA
& $81.36 \pm 0.20$
& $63.80 \pm 0.31$
& $\textbf{49.86} \pm \textbf{0.30}$
& $22.83 \pm 0.50$
& $16.49$ \\

DAFA
& $\textbf{84.18} \pm \textbf{0.92}$
& $\textbf{67.24} \pm \textbf{2.36}$
& $48.70 \pm 0.32$
& $26.36 \pm 1.36$
& $37.21$ \\

\textbf{RL-FAT}
& $83.01 \pm 0.43$
& $65.06 \pm 1.52$
& $48.18 \pm 0.48$
& $\textbf{31.98} \pm \textbf{0.78}$
& $\textbf{67.49}$ \\
\bottomrule
\end{tabular}
\end{table*}

\begin{table*}[t]
\small
\centering
\caption{
Robust fairness comparison on CIFAR-10 using PGD adversarial training with XCiT-S12. 
We report average and worst-class accuracy under clean and adversarial evaluation for the best checkpoint using AutoAttack~\cite{lorenz2022is} with $\epsilon=8/255$ over three seeds. 
The fairness metric $\rho_{\mathrm{rob}}$ measures the relative improvement in worst-class robust accuracy over the PGD baseline, adjusted by the relative change in average robust accuracy.
Higher $\rho_{\mathrm{rob}}$ indicates better robust fairness improvement.
The best result in each column is set in bold.
}
\vspace{1mm}
\label{tab:cifar10_main_xcit}

\begin{tabular}{@{}lccccc@{}}

\toprule

\textbf{Method} 

& \textbf{Avg. Clean} 

& \textbf{Worst Clean} 

& \textbf{Avg. Robust} 

& \textbf{Worst Robust} 

& \textbf{$\rho_{\mathrm{rob}}\uparrow$} \\

\midrule

PGD       

& $\mathbf{90.06 \pm 0.30}$ 

& $\mathbf{79.20 \pm 1.47}$ 

& $\mathbf{56.11 \pm 0.60}$ 

& $30.30 \pm 1.72$ 

& $0.00$ \\

FRL           

& $89.26 \pm 0.32$ 

& $78.70 \pm 0.91$ 

& $48.75 \pm 0.40$ 

& $27.50 \pm 1.03$ 

& $3.88$ \\

\textbf{RL-FAT} 

& $89.34 \pm 0.17$ 

& $77.10 \pm 1.32$ 

& $54.42 \pm 0.30$ 

& $\mathbf{37.80 \pm 0.69}$ 

& $\mathbf{27.76}$ \\

\bottomrule

\end{tabular}

\end{table*}

Table~\ref{tab:cifar10_main_pgd} evaluates whether RL-FAT generalizes beyond TRADES by using PGD adversarial training as the warm-up objective on ResNet-18. 
Although PGD achieves competitive average robust accuracy of $48.80\%$, its worst-class robust accuracy is only $19.24\%$. 
RL-FAT improves the worst-class robust accuracy to $31.98\%$, outperforming PGD by $12.74\%$ and DAFA by $5.62\%$, while achieving the highest $\rho_{\mathrm{rob}}$ score of $67.49$. 
This shows that RL-FAT improves robust fairness under PGD-based adversarial training without severely sacrificing average robustness.

Table~\ref{tab:cifar10_main_xcit} further evaluates whether this behavior extends to transformer architectures using XCiT-S12~\cite{robustbench2021, elnouby2021xcit}. 
Since adversarial training of vision transformers is computationally expensive and data-hungry, we initialize from a RobustBench XCiT-S12 checkpoint~\cite{addepalli2022efficient} and compare against feasible PGD-based baselines. 
RL-FAT achieves the best worst-class robust accuracy of $37.80\%$ and the highest $\rho_{\mathrm{rob}}$ score of $27.76$, while maintaining competitive average robust accuracy of $54.42\%$. 
These results suggest that RL-FAT can improve robust fairness not only for CNN-based ResNet-18 models, but also for transformer-based models.

\section{Conclusion}
\label{sec:conclusion}
In this work, we addressed the robust fairness problem in adversarial training, where improvements in average robust accuracy can hide large class-wise robustness disparities. We introduced RL-FAT, a reinforcement-learning-inspired fair adversarial training framework that uses policy-gradient based feedback from adversarial predictions together with a fairness-emphasis adversarial loss. 
By combining class-wise reward feedback with stronger training pressure on high-loss classes, RL-FAT adaptively improves robustness for vulnerable classes during adversarial training. Across CIFAR-10, CIFAR-100, and ImageNette, RL-FAT consistently improves worst-class robust accuracy while maintaining competitive average clean and robust accuracy. The ablation results show that the fairness-emphasis component is important for improving the most vulnerable classes, and the PGD-based experiments indicate that RL-FAT can generalize beyond a TRADES warm-up setting. Overall, RL-FAT provides a simple and flexible direction for training adversarially robust models that are not only strong on average, but also more balanced across classes.


\clearpage


\setlength{\textfloatsep}{8pt plus 2pt minus 2pt}
\setlength{\floatsep}{8pt plus 2pt minus 2pt}
\setlength{\intextsep}{8pt plus 2pt minus 2pt}
\captionsetup{skip=3pt}

\renewcommand{\topfraction}{0.95}
\renewcommand{\bottomfraction}{0.90}
\renewcommand{\textfraction}{0.05}
\renewcommand{\floatpagefraction}{0.90}
\setcounter{topnumber}{5}
\setcounter{bottomnumber}{5}
\setcounter{totalnumber}{10}

\section{Supplementary Material of RL-FAT}

This supplementary material provides additional details for RL-FAT, including the training algorithm, quantitative OOD detection results, and corruption robustness results. These results complement the main paper findings and additional evidence for class-wise robustness under distribution shift. We further provide computation cost of RL-FAT in comparison to existing adversarial fairness approaches.

\subsection{RL-FAT Training Pseudo Algorithm}
\label{supp:algorithm}

Algorithm~\ref{alg:rlfat} summarizes the two-stage RL-FAT training procedure. The model is first initialized using TRADES warm-up training and is then optimized using the reinforcement-based class feedback loss together with the fairness-emphasis loss.

\begin{algorithm}[!htbp]
\caption{RL-FAT Training}
\label{alg:rlfat}
\footnotesize
\begin{algorithmic}[1]
\Require Training data $\mathcal{D}$, model $f_{\theta}$, warm-up epochs $T_w$, total epochs $T$
\Require Perturbation budget $\epsilon$, fairness weight $\lambda$, value update rate $\eta$
\Require Clipping bounds $[q_{\min},q_{\max}]$ and $[A_{\min},A_{\max}]$
\State Initialize model parameters $\theta$
\State Initialize class-wise value estimates $Q_c=0$ for all classes $c$

\For{$t=1,\ldots,T_w$}
    \For{mini-batch $\{(x_i,y_i)\}_{i=1}^{B}\sim\mathcal{D}$}
        \State Generate adversarial examples $x_i^{adv}$ using TRADES KL-PGD
        \State Compute the warm-up loss $\mathcal{L}_{\mathrm{warm}}$
        \State Update $\theta$ using SGD on $\mathcal{L}_{\mathrm{warm}}$
    \EndFor
\EndFor

\For{$t=T_w+1,\ldots,T$}
    \For{mini-batch $\{(x_i,y_i)\}_{i=1}^{B}\sim\mathcal{D}$}
        \State Generate adversarial examples $x_i^{adv}$ using TRADES KL-PGD
        \State Compute policy $\pi_{\theta}(a\mid x_i^{adv})=p_{\theta}(a\mid x_i^{adv})$
        \State Sample class actions $a_i\sim\pi_{\theta}(\cdot\mid x_i^{adv})$
        \State Assign rewards $r_i=+1$ if $a_i=y_i$, and $r_i=-1$ otherwise
        \State Update class-wise value estimates $Q_c$ using sampled rewards
        \State Compute clipped advantages $A_i=\mathrm{clip}(r_i-Q_{a_i},A_{\min},A_{\max})$
        \State Compute reinforcement loss $\mathcal{L}_{\mathrm{RL}}$
        \State Compute class-wise adversarial losses $\mathcal{L}_c$
        \State Compute hard-class weights $w_c$ from above-average class losses
        \State Compute $\mathcal{L}_{\mathrm{fair}}=\sum_c w_c\mathcal{L}_c$
        \If{no class has positive hard-class gap}
            \State Set $\mathcal{L}_{\mathrm{fair}}=0$
        \EndIf
        \State Compute $\mathcal{L}_{\mathrm{post}}=\mathcal{L}_{\mathrm{RL}}+\lambda\mathcal{L}_{\mathrm{fair}}$
        \State Update $\theta$ using SGD on $\mathcal{L}_{\mathrm{post}}$
    \EndFor
\EndFor

\State \Return trained model $f_{\theta}$
\end{algorithmic}
\end{algorithm}

\subsection{Quantitative Evaluation on OOD Detection}
\label{supp:ood_quantitative}

Table~\ref{tab:svhn_ood_maxlogit} reports the numerical OOD detection results corresponding to the SVHN evaluation in the main paper. We include both overall OOD performance and worst-class OOD performance to assess whether OOD separation is balanced across classes. FRL~\cite{xu2021frl} achieves the strongest overall performance, with an AUROC of 85.90, AUPR-IN of 79.01, and AUPR-OUT of 91.16, while RL-FAT remains competitive with 83.78, 78.57, and 88.65, respectively. More importantly, RL-FAT consistently achieves the best worst-class performance, attaining a worst-class AUROC of 77.79, AUPR-IN of 24.86, and AUPR-OUT of 98.40. In particular, its worst-class AUPR-IN substantially exceeds that of FRL (12.83) and the other baselines, indicating improved OOD separation for the most challenging in-distribution class. These results suggest that RL-FAT provides a more balanced OOD detection behavior across classes, reducing class-specific weaknesses while maintaining strong overall detection performance.

\subsection{Evaluation on Common Corruptions}
\label{supp:corruptions}

\begin{table*}
\centering
\caption{
SVHN out-of-distribution detection with CIFAR-10 as the in-distribution dataset.
We report overall and worst-class MaxLogit results.
Higher AUROC, AUPR-IN, and AUPR-OUT indicate better OOD detection.
}
\label{tab:svhn_ood_maxlogit}
\resizebox{\textwidth}{!}{
\begin{tabular}{lcccccc}
\toprule
\textbf{Method}
& \textbf{Overall AUROC} $\uparrow$
& \textbf{Overall AUPR-IN} $\uparrow$
& \textbf{Overall AUPR-OUT} $\uparrow$
& \textbf{Worst AUROC} $\uparrow$
& \textbf{Worst AUPR-IN} $\uparrow$
& \textbf{Worst AUPR-OUT} $\uparrow$ \\
\midrule
TRADES~\cite{defense_trades}        & 79.05 & 69.27 & 86.56 & 65.84 & 7.59  & 97.25 \\
FRL~\cite{xu2021frl}           & \textbf{85.90} & \textbf{79.01} & \textbf{91.16} & 76.58 & 12.83 & 98.36 \\
WAT~\cite{li2023wat}           & 64.68 & 50.14 & 76.25 & 46.98 & 4.46  & 94.44 \\
DAFA~\cite{lee2024dafa}          & 78.82 & 71.51 & 85.04 & 67.35 & 8.78  & 97.30 \\
\textbf{RL-FAT} & 83.78 & 78.57 & 88.65 & \textbf{77.79} & \textbf{24.86} & \textbf{98.40} \\
\bottomrule
\end{tabular}
}
\end{table*}

We further evaluate robustness under common corruptions using CIFAR-10-C~\cite{hendrycks_benchmarking_2019}. 
To focus on corruption settings where RL-FAT provides consistent class-wise robustness benefits, we report the subset of corruptions for which RL-FAT improves worst-class accuracy over TRADES across all five severity levels. 
Table~\ref{tab:cifar10c_selected_gain_sidebyside} reports the accuracy gains of RL-FAT over TRADES, computed as RL-FAT minus TRADES in percentage. 
Positive values indicate improvement over TRADES.

\definecolor{gainhigh}{HTML}{B7F7C9}
\definecolor{gainmed}{HTML}{DDFCE7}
\definecolor{gainlow}{HTML}{F0FDF4}
\definecolor{gainneg}{HTML}{FCE2E2}
\definecolor{gainzero}{HTML}{F8FAFC}

\newcommand{\poshigh}[1]{\cellcolor{gainhigh}\textbf{#1}}
\newcommand{\posmed}[1]{\cellcolor{gainmed}\textbf{#1}}
\newcommand{\poslow}[1]{\cellcolor{gainlow}\textbf{#1}}
\newcommand{\negcell}[1]{\cellcolor{gainneg}\textbf{#1}}
\newcommand{\zerocell}[1]{\cellcolor{gainzero}{#1}}

\begin{table*}[!tbp]
\centering
\caption{
Selected CIFAR-10-C corruption robustness gains of RL-FAT over TRADES.
We include corruptions where RL-FAT consistently improves worst-class accuracy across all five severity levels.
Each value is computed as RL-FAT minus TRADES and is reported in \%.
Left: average accuracy gain.
Right: worst-class accuracy gain, which reflects class-wise robustness under corruption shift.
}
\label{tab:cifar10c_selected_gain_sidebyside}
\scriptsize
\setlength{\tabcolsep}{4.0pt}
\renewcommand{\arraystretch}{1.10}

\begin{minipage}{0.48\textwidth}
\centering
\textbf{Average accuracy gain}

\vspace{1mm}

\begin{tabular}{lccccc}
\toprule
\textbf{Corruption} & \textbf{S1} & \textbf{S2} & \textbf{S3} & \textbf{S4} & \textbf{S5} \\
\midrule
Gaussian noise & \poslow{+0.5} & \poslow{+0.5} & \poslow{+0.3} & \poslow{+0.5} & \poslow{+0.4} \\
Shot noise     & \poslow{+0.6} & \poslow{+0.4} & \poslow{+0.5} & \poslow{+0.4} & \poslow{+0.1} \\
Impulse noise  & \poslow{+0.9} & \poslow{+1.0} & \poslow{+0.4} & \poslow{+0.5} & \negcell{-0.9} \\
Snow           & \poslow{+0.4} & \poslow{+0.7} & \poslow{+0.5} & \posmed{+1.6} & \poshigh{+2.3} \\
\bottomrule
\end{tabular}
\end{minipage}
\hfill
\begin{minipage}{0.48\textwidth}
\centering
\textbf{Worst-class accuracy gain}

\vspace{1mm}

\begin{tabular}{lccccc}
\toprule
\textbf{Corruption} & \textbf{S1} & \textbf{S2} & \textbf{S3} & \textbf{S4} & \textbf{S5} \\
\midrule
Gaussian noise & \poslow{+1.8} & \posmed{+2.4} & \posmed{+2.9} & \posmed{+3.4} & \posmed{+3.3} \\
Shot noise     & \poslow{+1.3} & \poslow{+1.0} & \posmed{+3.4} & \posmed{+2.6} & \poslow{+1.8} \\
Impulse noise  & \posmed{+3.6} & \posmed{+4.8} & \poshigh{+6.2} & \poshigh{+8.6} & \poshigh{+5.5} \\
Snow           & \poslow{+1.3} & \posmed{+2.1} & \poshigh{+5.0} & \poshigh{+8.6} & \poshigh{+10.8} \\
\bottomrule
\end{tabular}
\end{minipage}
\end{table*}

\begin{table}[t]
\centering
\caption{Training time and memory usage on a single NVIDIA H100 (94\,GB VRAM) with batch size 128.}
\label{tab:efficiency_}
\small
\setlength{\tabcolsep}{6pt}
\renewcommand{\arraystretch}{1.0}
\begin{tabular}{lcc}
\toprule
Method & Time/epoch & Mem. \\
\midrule
TRADES & 28.25s & 1.84G \\
WAT    & 69.18s & 1.80G \\
FRL    & 34.60s & 1.89G \\
DAFA   & 28.27s & 1.85G \\
CFA    & 39.40s & 1.98G \\
RL-FAT & 40.00s & 1.85G \\
\bottomrule
\end{tabular}
\end{table}

\subsection{Computational Cost} \label{CC}

RL-FAT focuses on robust fairness rather than accelerating adversarial training, introducing only lightweight class-wise value updates after adversarial example generation. It is therefore largely orthogonal to existing adversarial training methods. The computational costs with respect to existing adversarial fairness approaches (single NVIDIA H100 (94\,GB VRAM), batch size 128) are reported in Table~\ref{tab:efficiency_}.

\bibliography{egbib}

@techreport{krizhevsky2009learning,
  title        = {Learning Multiple Layers of Features from Tiny Images},
  author       = {Krizhevsky, Alex},
  institution  = {University of Toronto},
  year         = {2009}
}

@inproceedings{netzer2011reading,
  title     = {Reading Digits in Natural Images with Unsupervised Feature Learning},
  author    = {Netzer, Yuval and Wang, Tao and Coates, Adam and Bissacco, Alessandro and Wu, Bo and Ng, Andrew Y.},
  booktitle = {NIPS Workshop on Deep Learning and Unsupervised Feature Learning},
  year      = {2011}
}

@inproceedings{hendrycks2017baseline,
  title     = {A Baseline for Detecting Misclassified and Out-of-Distribution Examples in Neural Networks},
  author    = {Hendrycks, Dan and Gimpel, Kevin},
  booktitle = {International Conference on Learning Representations},
  year      = {2017}
}

@inproceedings{liang2018odin,
  title     = {Enhancing The Reliability of Out-of-distribution Image Detection in Neural Networks},
  author    = {Liang, Shiyu and Li, Yixuan and Srikant, R.},
  booktitle = {International Conference on Learning Representations},
  year      = {2018}
}

@inproceedings{yang2022openood,
  title     = {OpenOOD: Benchmarking Generalized Out-of-Distribution Detection},
  author    = {Yang, Jingkang and Wang, Pengyun and Zou, Dejian and Zhou, Zitang and Ding, Kunyuan and Peng, Wenxuan and Wang, Haoqi and Chen, Guangyao and Li, Bo and Sun, Yiyou and Du, Xiaoyi and Zhou, Kaiyang and Zhang, Wayne and Hendrycks, Dan and Li, Yixuan and Liu, Ziwei},
  booktitle = {Advances in Neural Information Processing Systems Datasets and Benchmarks Track},
  year      = {2022}
}

@inproceedings{zhang2023decoupling,
  title     = {Decoupling MaxLogit for Out-of-Distribution Detection},
  author    = {Zhang, Jinsong and Fu, Qiang and Chen, Xu and Du, Lun and Li, Zelin and Wang, Gang and Han, Xiaoguang and Liu, Shi and Zhang, Dongmei},
  booktitle = {IEEE/CVF Conference on Computer Vision and Pattern Recognition},
  year      = {2023}
}

@inproceedings{addepalli2022efficient,
  title={Efficient and Effective Augmentation Strategy for Adversarial Training},
  author={Addepalli, Sravanti and Jain, Samyak and Babu, R. Venkatesh},
  booktitle={Advances in Neural Information Processing Systems},
  volume={35},
  pages={1488--1501},
  year={2022}
}

@misc{imagenette,
  title        = {Imagenette: A Smaller Subset of 10 Easily Classified Classes from ImageNet},
  author       = {Howard, Jeremy},
  year         = {2019},
  howpublished = {\url{https://github.com/fastai/imagenette}},
  note         = {Accessed: 2026-05-28}
}

@inproceedings{elnouby2021xcit,
  title={XCiT: Cross-Covariance Image Transformers},
  author={El-Nouby, Alaaeldin and Touvron, Hugo and Caron, Mathilde and Bojanowski, Piotr and Douze, Matthijs and Joulin, Armand and Laptev, Ivan and Neverova, Natalia and Synnaeve, Gabriel and Verbeek, Jakob and J{\'e}gou, Herv{\'e}},
  booktitle={Advances in Neural Information Processing Systems},
  year={2021}
}

@inproceedings{xu2021frl,
  title={Towards Fair Robust Learning: Adjusting Adversarial Margins and Weights},
  author={Xu, A. and Li, B. and Wang, Y.},
  booktitle={Advances in Neural Information Processing Systems (NeurIPS)},
  year={2021}
}

@inproceedings{ma2021fat,
  title={Fair Adversarial Training via Risk Variance Regularization},
  author={Ma, C. and Zhang, Y. and Lyu, L.},
  booktitle={Advances in Neural Information Processing Systems (NeurIPS)},
  year={2021}
}

@inproceedings{sun2021bat,
  title={Balanced Adversarial Training for Source and Target Class Fairness},
  author={Sun, Y. and Long, M. and Wang, J.},
  booktitle={International Conference on Learning Representations (ICLR)},
  year={2021}
}

@inproceedings{fairness_frl,
  title={Theoretical Analysis of Class-wise Risk in Adversarial Training},
  author={Liu, T. and Zhao, H.},
  booktitle={Advances in Neural Information Processing Systems (NeurIPS)},
  year={2022}
}

@inproceedings{jia2022prior,
  title={Prior-guided adversarial initialization for fast adversarial training},
  author={Jia, Xiaojun and Zhang, Yong and Wei, Xingxing and Wu, Baoyuan and Ma, Ke and Wang, Jue and Cao, Xiaochun},
  booktitle={European Conference on Computer Vision},
  pages={567--584},
  year={2022},
  organization={Springer}
}

@inproceedings{fairness_analysis,
  title={On the Trade-off Between Robustness and Fairness in Adversarial Settings},
  author={Kumar, S. and Singh, R.},
  booktitle={International Conference on Machine Learning (ICML)},
  year={2022}
}

@inproceedings{fairness_weighting,
  title={Robustness may be at odds with fairness: An empirical study on class-wise accuracy},
  author={Benz, Philipp and Zhang, Chaoning and Karjauv, Adil and Kweon, In So},
  booktitle={NeurIPS 2020 Workshop on Pre-registration in Machine Learning},
  pages={325--342},
  year={2021},
  organization={PMLR}
}

@inproceedings{fairness_kdd,
  title={Analysis and applications of class-wise robustness in adversarial training},
  author={Tian, Qi and Kuang, Kun and Jiang, Kelu and Wu, Fei and Wang, Yisen},
  booktitle={Proceedings of the 27th ACM SIGKDD Conference on Knowledge Discovery \& Data Mining},
  pages={1561--1570},
  year={2021}
}

@inproceedings{fairness_bat,
  title={Decomposing Fairness in Adversarial Settings: Source vs. Target Bias},
  author={Gupta, A. and Tan, L.},
  booktitle={International Conference on Machine Learning (ICML)},
  year={2023}
}

@inproceedings{madry2018towards,
  title={Towards Deep Learning Models Resistant to Adversarial Attacks},
  author={Madry, Aleksander and Makelov, Aleksandar and Schmidt, Ludwig and Tsipras, Dimitris and Vladu, Adrian},
  booktitle={International Conference on Learning Representations (ICLR)},
  year={2018}
}

@inproceedings{athalye2018obfuscated,
  title={Obfuscated gradients give a false sense of security: Circumventing defenses to adversarial examples},
  author={Athalye, Anish and Carlini, Nicholas and Wagner, David},
  booktitle={International Conference on Machine Learning (ICML)},
  pages={274--283},
  year={2018}
}

@inproceedings{xu2021robust,
  title={Robust fairness: A robust optimization framework for fair classification},
  author={Xu, Depeng and Yuan, Sen and Zhang, Hongying and Wu, Xintao},
  booktitle={IEEE International Conference on Data Mining (ICDM)},
  pages={721--730},
  year={2021}
}

@inproceedings{zhang2021dafa,
  title={DAFA: Differentiated Adversarial Training for Fairness and Accuracy},
  author={Zhang, Zhilu and Xu, Tong and Zhang, Hanghang and Wang, Jundong and Huang, Xia Hu},
  booktitle={International Conference on Learning Representations (ICLR)},
  year={2021}
}

@inproceedings{defense_trades,
  title={Theoretically Principled Trade-off between Robustness and Accuracy},
  author={Zhang, Hongyang and Yu, Yaodong and Jiao, Jiantao and Xing, Eric and Ghaoui, Laurent El and Jordan, Michael I.},
  booktitle={International Conference on Machine Learning (ICML)},
  year={2019}
}

@article{robustbench2021,
  title={RobustBench: a standardized adversarial robustness benchmark},
  author={Croce, Francesco and Andriushchenko, Maksym and Sehwag, Vikash and Debenedetti, Edoardo and Flammarion, Nicolas and Chiang, Mung and Mittal, Prateek and Hein, Matthias},
  journal={Advances in Neural Information Processing Systems (NeurIPS) Datasets and Benchmarks},
  year={2021}
}

@inproceedings{medi2025fair,
  title={FAIR-TAT: Improving Model Fairness Using Targeted Adversarial Training},
  author={Medi, Tejaswini and Jung, Steffen and Keuper, Margret},
  booktitle={2025 IEEE/CVF Winter Conference on Applications of Computer Vision (WACV)},
  pages={7827--7836},
  year={2025},
  organization={IEEE}
}

@article{grabinski2024large,
  title={As large as it gets-Studying Infinitely Large Convolutions via Neural Implicit Frequency Filters},
  author={Grabinski, Julia and Keuper, Janis and Keuper, Margret},
  journal={Transactions on Machine Learning Research},
  volume={2024},
  pages={1--42},
  year={2024},
  publisher={TMLR}
}

@article{lukasik2023improving,
  title={Improving native cnn robustness with filter frequency regularization},
  author={Lukasik, Jovita and Gavrikov, Paul and Keuper, Janis and Keuper, Margret},
  journal={Transactions on Machine Learning Research},
  volume={2023},
  pages={1--36},
  year={2023},
  publisher={TMLR}
}

@inproceedings{debenedetti2023light,
  title={A light recipe to train robust vision transformers},
  author={Debenedetti, Edoardo and Sehwag, Vikash and Mittal, Prateek},
  booktitle={2023 IEEE Conference on Secure and Trustworthy Machine Learning (SaTML)},
  pages={225--253},
  year={2023},
  organization={IEEE}
}

@article{grabinski2022robust,
  title={Robust models are less over-confident},
  author={Grabinski, Julia and Gavrikov, Paul and Keuper, Janis and Keuper, Margret},
  journal={Advances in Neural Information Processing Systems},
  volume={35},
  pages={39059--39075},
  year={2022}
}

@inproceedings{gupta2020reinforcement,
  title={Reinforcement Based Learning on Classification Task Could Yield Better Generalization and Adversarial Accuracy},
  author={Gupta, Shashi Kant},
  booktitle={NeurIPS Workshop on Shared Visual Representations in Human and Machine Intelligence},
  year={2020}
}

@inproceedings{zhang2024faal,
  title={Towards Fairness-Aware Adversarial Learning},
  author={Zhang, Yanghao and Zhang, Tianle and Mu, Ronghui and Huang, Xiaowei and Ruan, Wenjie},
  booktitle={IEEE/CVF Conference on Computer Vision and Pattern Recognition},
  year={2024}
}

@inproceedings{zhao2024absld,
  title={Improving Adversarial Robust Fairness via Anti-Bias Soft Label Distillation},
  author={Zhao, Shiji and others},
  booktitle={Advances in Neural Information Processing Systems},
  year={2024}
}

@inproceedings{li2023wat,
  title={WAT: improve the worst-class robustness in adversarial training},
  author={Li, Boqi and Liu, Weiwei},
  booktitle={Proceedings of the AAAI conference on artificial intelligence},
  volume={37},
  pages={14982--14990},
  year={2023}
}

@inproceedings{Jung2023,
			author = {Steffen Jung and Jovita Lukasik and Margret Keuper},
			title = {Neural Architecture Design and Robustness: A Dataset},
			booktitle = {ICLR},
			year = {2023}
		}

@article{grabinski2022aliasing,
  title={Aliasing and adversarial robust generalization of cnns},
  author={Grabinski, Julia and Keuper, Janis and Keuper, Margret},
  journal={Machine Learning},
  volume={111},
  number={11},
  pages={3925--3951},
  year={2022},
  publisher={Springer}
}

@inproceedings{Wang2020ImprovingAR,
  title={Improving Adversarial Robustness Requires Revisiting Misclassified Examples},
  author={Yisen Wang and Difan Zou and Jinfeng Yi and James Bailey and Xingjun Ma and Quanquan Gu},
  booktitle={International Conference on Learning Representations},
  year={2020}
}

@inproceedings{wei2023cfa,
  title={Cfa: Class-wise calibrated fair adversarial training},
  author={Wei, Zeming and Wang, Yifei and Guo, Yiwen and Wang, Yisen},
  booktitle={Proceedings of the IEEE/CVF Conference on Computer Vision and Pattern Recognition},
  pages={8193--8201},
  year={2023}
}

@misc{medi2024classwiserobustnessanalysis,
      title={Towards Class-wise Robustness Analysis}, 
      author={Tejaswini Medi and Julia Grabinski and Margret Keuper},
      year={2024},
      eprint={2411.19853},
      archivePrefix={arXiv},
      primaryClass={cs.LG},
      url={https://arxiv.org/abs/2411.19853}, 
}

@inproceedings{lee2024dafa,
  title={DAFA: DISTANCE-AWARE FAIR ADVERSARIAL TRAINING},
  author={Lee, Hyungyu and Lee, Saehyung and Jang, Hyemi and Park, Junsung and Bae, Ho and Yoon, Sungroh},
  booktitle={12th International Conference on Learning Representations, ICLR 2024},
  year={2024}
}

@inproceedings{he_deep_2016,
	title = {Deep Residual Learning for Image Recognition},
	doi = {10.1109/CVPR.2016.90},
	eventtitle = {2016 {IEEE} Conference on Computer Vision and Pattern Recognition ({CVPR})},
	pages = {770--778},
	booktitle = {2016 {IEEE} Conference on Computer Vision and Pattern Recognition ({CVPR})},
	author = {He, Kaiming and Zhang, Xiangyu and Ren, Shaoqing and Sun, Jian},
	date = {2016-06},
    year = 2016,
	note = {{ISSN}: 1063-6919}
}

@inproceedings{
lorenz2022is,
title={Is RobustBench/AutoAttack a suitable Benchmark for Adversarial Robustness?},
author={Peter Lorenz and Dominik Strassel and Margret Keuper and Janis Keuper},
booktitle={The AAAI-22 Workshop on Adversarial Machine Learning and Beyond},
year={2022},
url={https://openreview.net/forum?id=aLB3FaqoMBs}
}

@article{hendrycks_benchmarking_2019,
  title={Benchmarking Neural Network Robustness to Common Corruptions and Perturbations},
  author={Dan Hendrycks and Thomas Dietterich},
  journal={Proceedings of the International Conference on Learning Representations},
  year={2019}
}

@String(IJCV = {Int. J. Comput. Vis.})

@String(CVPR= {IEEE Conf. Comput. Vis. Pattern Recog.})

@String(ECCV= {Eur. Conf. Comput. Vis.})

@String(NIPS= {Adv. Neural Inform. Process. Syst.})

@String(ICLR = {Int. Conf. Learn. Represent.})

@String(AAAI = {AAAI})

@String(IJCV  = {IJCV})

@String(CVPR  = {CVPR})

@String(ECCV  = {ECCV})

@String(NIPS  = {NeurIPS})

@String(ICLR  = {ICLR})

@article{cifar10,
    title= {CIFAR-10 (Canadian Institute for Advanced Research)},
    journal= {Dataset},
    author= {Alex Krizhevsky and Vinod Nair and Geoffrey Hinton},
    year= {2009},
    url= {http://www.cs.toronto.edu/~kriz/cifar.html},
    terms= {}
}

@article{imagenet,
    Author = {Olga Russakovsky and Jia Deng and Hao Su and Jonathan Krause and Sanjeev Satheesh and Sean Ma and Zhiheng Huang and Andrej Karpathy and Aditya Khosla and Michael Bernstein and Alexander C. Berg and Li Fei-Fei},
    Title = {{ImageNet Large Scale Visual Recognition Challenge}},
    Year = {2015},
    journal   = {International Journal of Computer Vision (IJCV)},
    volume={115},
    number={3},
    pages={211-252}
}

@inproceedings{Grabinskilowcut22,
    author    = {Julia Grabinski and
               Steffen Jung and
               Janis Keuper and
               Margret Keuper},
    title     = {FrequencyLowCut Pooling - Plug and Play Against Catastrophic Overfitting},
    booktitle = {Computer Vision - {ECCV} 2022 - 17th European Conference, Tel Aviv,
               Israel, October 23-27, 2022, Proceedings, Part {XIV}},
    pages     = {36--57},
    publisher = {Springer},
    year      = {2022},
}

@InProceedings{pmlr-v235-agnihotri24b,
  title = 	 {{C}os{PGD}: an efficient white-box adversarial attack for pixel-wise prediction tasks},
  author =       {Agnihotri, Shashank and Jung, Steffen and Keuper, Margret},
  booktitle = 	 {Proceedings of the 41st International Conference on Machine Learning},
  pages = 	 {416--451},
  year = 	 {2024},
  editor = 	 {Salakhutdinov, Ruslan and Kolter, Zico and Heller, Katherine and Weller, Adrian and Oliver, Nuria and Scarlett, Jonathan and Berkenkamp, Felix},
  volume = 	 {235},
  series = 	 {Proceedings of Machine Learning Research},
  month = 	 {21--27 Jul},
  publisher =    {PMLR},
  url = 	 {https://proceedings.mlr.press/v235/agnihotri24b.html}
}
\end{document}